\documentclass[10pt,journal,compsoc]{IEEEtran}

\usepackage{stfloats}
\usepackage{footnote}
\usepackage[accsupp]{axessibility}
\usepackage{graphicx}
\usepackage{amsmath}
\usepackage{amssymb}
\usepackage{booktabs}
\usepackage{amsfonts,amssymb}

\usepackage{hyperref}
\hypersetup{pagebackref,breaklinks,colorlinks}
\usepackage{soul}
\usepackage{bbm}
\usepackage{colortbl}

\usepackage[capitalize]{cleveref}
\crefname{section}{Sec.}{Secs.}
\Crefname{section}{Section}{Sections}
\Crefname{table}{Table}{Tables}
\crefname{table}{Tab.}{Tabs.}

\newcommand{\etal}{{\textit{et al.}}}
\begin{document}

\title{DN-DETR: Accelerate DETR Training by Introducing Query DeNoising}

% \author{Feng Li$^{1,2}$\thanks{Indicates equal contribution.} \thanks{This work was done when Feng Li and Hao Zhang were interns at IDEA. }, ~Hao Zhang$^{1,2*\dag}$, ~Shilong Liu$^{2,3}$ , ~Jian Guo$^{2}$,
%  ~Lionel M. Ni$^{1,4}$, ~Lei Zhang$^{2}$\thanks{Corresponding author.} \\
% $^1$The Hong Kong University of Science and Technology. \\
% $^2$International Digital Economy Academy (IDEA). \\
% $^3$Tsinghua University. \\
% % $^3$Dept. of Comp. Sci. and Tech., BNRist Center, State Key Lab for Intell. Tech. \\
% % \& Sys., Institute for AI, Tsinghua-Bosch Joint Center for ML, Tsinghua University. \\
% $^4$The Hong Kong University of Science and Technology (Guangzhou).\\
% \texttt{\{fliay,hzhangcx\}@connect.ust.hk} \\
% \texttt{\{liusl20\}@mails.tsinghua.edu.cn} \\
% \texttt{\{ni\}@ust.hk} \\
% \texttt{\{guojian,leizhang\}@idea.edu.cn} \\
% }

\author{Feng~Li${^*}$, Hao~Zhang${^*}$, Shilong~Liu, Jian Guo, Lionel~M.~Ni,
        and~Lei~Zhang,~\IEEEmembership{IEEE~Fellow}% <-this % stops a space
\IEEEcompsocitemizethanks{\IEEEcompsocthanksitem Feng Li and Hao Zhang are with the Department
of Computer Science and Engineering, The Hong Kong University of Science and Technology, Hong Kong.

\IEEEcompsocthanksitem Shilong Liu is with the Department
of Computer Science and Engineering, Tsinghua University, Beijing.

\IEEEcompsocthanksitem Lionel Ni is the president of The Hong Kong University of Science and Technology (Guangzhou).
\IEEEcompsocthanksitem Jian Guo and Lei Zhang are with IDEA. 

\IEEEcompsocthanksitem ${^*}$ denotes equal contribution.}% <-this % stops an unwanted space
% \thanks{Manuscript received April 19, 2005; revised August 26, 2015.}

}

% \markboth{Journal of \LaTeX\ Class Files,~Vol.~14, No.~8, August~2015}%
% {Shell \MakeLowercase{\textit{et al.}}: Bare Demo of IEEEtran.cls for Computer Society Journals}

\markboth{DN-DETR: Accelerate DETR Training by Introducing Query DeNoising}%
{Shell \MakeLowercase{\textit{et al.}}: Bare Demo of IEEEtran.cls for Computer Society Journals}

\IEEEtitleabstractindextext{%
\begin{abstract}
We present in this paper a novel denoising training method to speed up DETR (DEtection TRansformer) training and offer a deepened understanding of the slow convergence issue of DETR-like methods. We show that the slow convergence results from the instability of bipartite graph matching which causes inconsistent optimization goals in early training stages. To address this issue, except for the Hungarian loss, our method additionally feeds GT bounding boxes with noises into the Transformer decoder and trains the model to reconstruct the original boxes, which effectively reduces the bipartite graph matching difficulty and leads to faster convergence. Our method is universal and can be easily plugged into any DETR-like method by adding dozens of lines of code to achieve a remarkable improvement. As a result, our DN-DETR results in a remarkable improvement ($+1.9$AP) under the same setting and achieves $46.0$ AP and $49.5$ AP trained for $12$ and $50$ epochs with the ResNet-$50$ backbone. Compared with the baseline under the same setting, DN-DETR achieves comparable performance with $50\%$ training epochs. We also demonstrate the effectiveness of denoising training in CNN-based detectors (Faster R-CNN), segmentation models (Mask2Former, Mask DINO), and more DETR-based models (DETR, Anchor DETR, Deformable DETR). Code is available at \url{https://github.com/IDEA-Research/DN-DETR}.
\end{abstract}

}
\maketitle

\IEEEdisplaynontitleabstractindextext

% \fi
\IEEEpeerreviewmaketitle

%%%%%%%%% ABSTRACT

%%%%%%%%% BODY TEXT
\IEEEraisesectionheading{\section{Introduction}}
\label{sec:intro}
Object detection is a fundamental task in computer vision that aims to predict the bounding boxes and classes of objects in an image. While having made remarkable progress, classical detectors ~\cite{ren2015faster, redmon2018yolov3} were mainly based on convolutional neural networks, until Carion \etal ~\cite{carion2020end} recently introduced Transformers ~\cite{vaswani2017attention} into object detection and proposed DETR (DEtection TRansformer). 

In contrast to previous detectors, DETR uses learnable queries to probe image features from the output of Transformer encoders and bipartite graph matching to perform set-based box prediction. Such a design effectively eliminates hand-designed anchors and non-maximum suppression (NMS) and makes object detection end-to-end optimizable. However, DETR suffers from prohibitively slow training convergence compared with previous detectors. To obtain a good performance, it usually takes $500$ epochs of training on the COCO detection dataset, in contrast to $12$ epochs used in the original Faster-RCNN training.

\begin{figure}
    \centering
    % \vspace{-0.2cm}
    \includegraphics[scale=0.4]{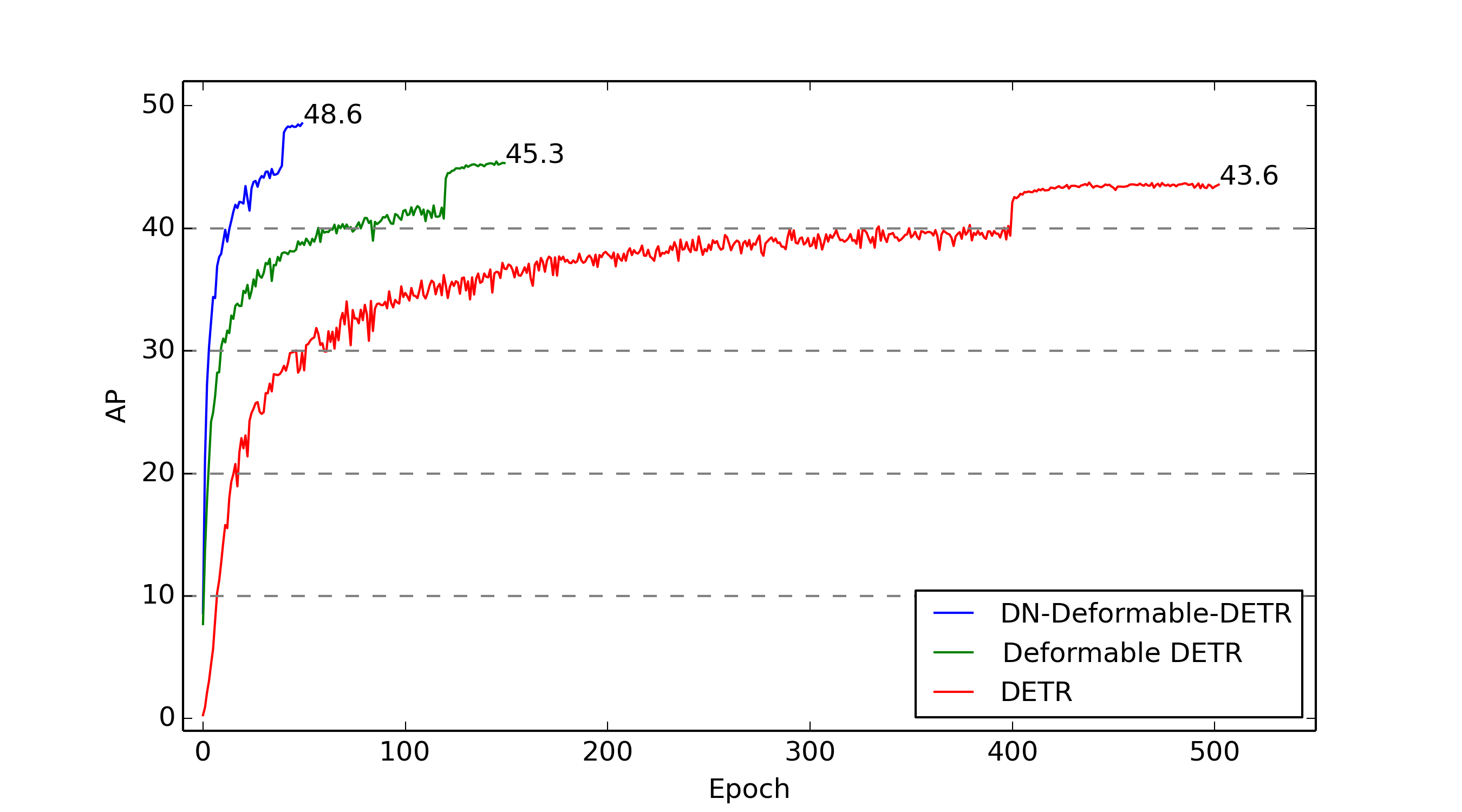}
    % \vspace{-0.3cm}
    \caption{Convergence curve between our model DN-Deformable-DETR built upon Deformable DETR with denoising training and previous models under ResNet-50 backbone. }
    % \vspace{-0.5cm}
    \label{fig:converge}
\end{figure}

Much work ~\cite{wang2021anchor,meng2021conditional,zhu2020deformable,sun2020rethinking,liu2022dabdetr,dai2021dynamic} has tried to identify the root cause and mitigate the slow convergence issue. Some of them address the problem by improving the model architecture. For example, Sun \etal ~\cite{sun2020rethinking} attributed the slow convergence issue to the low efficiency of the cross-attention and proposed an encoder-only DETR. Dai \etal ~\cite{dai2021dynamic} designed an RoI-based dynamic decoder to help the decoder focus on regions of interest. More recent works propose to associate each DETR query with a specific spatial position rather than multiple positions for more efficient feature probing ~\cite{wang2021anchor,meng2021conditional,zhu2020deformable,liu2022dabdetr}. For instance, Conditional DETR ~\cite{meng2021conditional} decouples each query into a content part and a positional part, enforcing a query to have a clear correspondence with a specific spatial position. Deformable DETR ~\cite{zhu2020deformable} and Anchor DETR ~\cite{wang2021anchor} directly treat $2D$ reference points as queries to perform cross-attention. DAB-DETR ~\cite{liu2022dabdetr} interprets queries as $4$-D anchor boxes and learns to progressively improve them layer by layer. 

Despite all the progress, few works pay attention to the bipartite graph matching part for more efficient training. In this study, we find that the slow convergence issue also results from the discrete bipartite graph matching component, which is unstable especially in the early stages of training due to the nature of stochastic optimization. As a consequence, for the same image, a query is often matched with different objects in different epochs, which makes optimization ambiguous and inconstant.

To address this problem, we propose a novel training method by introducing a query denoising task to help stabilize bipartite graph matching in the training process. Since previous works have shown effectiveness in interpreting queries as reference points ~\cite{zhu2020deformable, wang2021anchor} or anchor boxes ~\cite{liu2022dabdetr}, which contain positional information, we follow their viewpoint and use $4$D anchor boxes as queries. 
Our solution is to feed noised GT bounding boxes as noised queries together with learnable anchor queries into Transformer decoders. Both kinds of queries have the same input format of $(x,y,w,h)$ and can be fed into Transformer decoders simultaneously. For noised queries, we perform a denoising task to reconstruct their corresponding GT boxes. For other learnable anchor queries, we use the same training loss and bipartite matching as in the vanilla DETR. 
As the noised bounding boxes do not need to go through the bipartite graph matching component, the denoising task can be regarded as an easier auxiliary task, helping DETR alleviate the unstable discrete bipartite matching and learn bounding box prediction more quickly.
Meanwhile, the denoising task also helps lower the optimization difficulty because the added random noise is usually small.
To maximize the potential of this auxiliary task, we also regard each decoder query as a bounding box + a class label embedding so that we are able to conduct both box denoising and label denoising. 

In summary, our method is a denoising training approach. Our loss function consists of two components. One is a reconstruction loss and the other is a Hungarian loss which is the same as in other DETR-like methods. Our method can be easily plugged into any existing DETR-like method. For convenience, we utilize DAB-DETR ~\cite{liu2022dabdetr} to evaluate our method since their decoder queries are explicitly formulated as $4$D anchor boxes $(x, y, w, h)$. For DETR variants that only support $2$D anchor points such as anchor DETR ~\cite{wang2021anchor}, we can do denoising on anchor points. For those that do not support anchors like the vanilla DETR ~\cite{carion2020end}, we can do linear transformation to map $4$D anchor boxes to the same latent space as for other learnable queries. 

To the best of our knowledge, this is the first work to introduce the denoising principle into detection models. We summarize our contribution as follows:
\begin{enumerate}

\item We design a novel training method to speed up DETR training. Experimental results show that our method not only accelerates training convergence but also leads to a remarkably better training result --- achieving the best result among all detection algorithms in the $12$-epoch setting. Moreover, our method shows a remarkable improvement ($+\textbf{1.9}$ AP) over our baseline DAB-DETR and can be easily integrated into other DETR-like methods.

\item We analyze the slow convergence of DETR from a novel viewpoint and give a deeper understanding of DETR training. We design a metric to evaluate the instability of bipartite matching and verify that our method can effectively lower the instability.

\item We conduct a series of ablation studies to analyze the effectiveness of different components of our model, such as noise, label embedding, and attention mask.
\end{enumerate}

This paper is an extension of our previous paper \cite{li2022dn} that was accepted to CVPR'2022 as an oral presentation. Compared with its conference version, this paper brings some new contributions as follows.
\begin{enumerate}
\item We achieve better results and faster convergence by introducing deformable attention into our decoder layer.
\item We further demonstrate the effectiveness of denoising training by adding it to other DETR-like models without 4D anchor design, including Vanilla DETR without explicit anchors and Anchor DETR with only 2D anchors. We also show denoising training can improve segmentation models such as Mask2Former and Mask DINO.
\item We incorporate denoising training to the traditional CNN detector Faster R-CNN to show its generalization ability.
\item We provide more experimental results and analysis to get a better understanding of our method.
\end{enumerate}
\section{Related Work}
\subsection{Classical CNN Detectors}
Most modern object detection models are based on convolutional networks, which have achieved significant success in recent years. Classical CNN-based detectors can be divided into $2$ categories, one-stage, and two-stage methods. Two-stage methods like HTC~\cite{chen2019hybrid} and Fast R-CNN~\cite{gao2021fast} first generate some region proposals and then decide whether each region contains an object and do bounding box regression to get a refined box. Ren \etal ~\cite{ren2015faster} proposed an end-to-end method that utilizes a Region Proposal Network to predict anchor boxes. In contrast to two-stage methods, one-stage methods, including YOLO900~\cite{redmon2016yolo9000} and YOLOv3~\cite{redmon2018yolov3} directly predict the offset of real boxes relative to anchor boxes.

Though these methods achieve top performance on many datasets, they are sensitive to the way how anchors are generated. In addition, they require some hand-crafted components like non-maximum suppression (NMS) and label assignment rules. Therefore, they suffer from these drawbacks and can not be end-to-end optimized.
\subsection{DETR-based Detectors}
Carion \etal~\cite{carion2020end} proposed an end-to-end object detector based on Transformers ~\cite{vaswani2017attention} named DETR (DEtection TRansformer) without using anchors. While DETR achieves comparable results with Faster-RCNN \cite{ren2015faster}, its training suffers severely from the slow convergence problem --- it needs $500$ epochs of training to obtain a good performance.   

Many recent works have attempted to speed up the training process of DETR. Some find the cross attention of Transformer decoders in DETR inefficient and make improvements in different ways. For example, Dai \etal ~\cite{Dai_2021_ICCV} designed a dynamic decoder that can focus on regions of interest in a coarse-to-fine manner and lower the learning difficulty. Sun \etal ~\cite{sun2020rethinking} discarded the Transformer decoder and proposed an encoder-only DETR. Another series of works make improvements in decoder queries. Zhu \etal ~\cite{zhu2020deformable} designed an attention module that only attends to some sampling points around a reference point. Meng et al.~\cite{meng2021conditional} decoupled each decoder query into a content part and a position part and only utilized the content-to-content and position-to-position terms in the cross-attention formulation. Yao \etal ~\cite{yao2021efficient} utilized a Region Proposal Network (RPN) to propose top-$K$ anchor points. DAB-DETR ~\cite{liu2022dabdetr} uses $4$-D box coordinates as queries and updates boxes layer by layer in a cascade manner.

Despite all the progress, none of them treats bipartite graph matching used in the Hungarian loss as the main reason for slow convergence. Sun \etal ~\cite{sun2020rethinking} analyzed the impact of Hungarian loss by using a pre-trained DETR as a teacher to provide the GT label assignment for a student model and train the student model. They found that the label assignment only helps the convergence in the early stage of training but does not influence the final performance significantly. Therefore, they concluded that the Hungarian loss is not the main reason for the slow convergence. In this work, we give a different analysis with an effective solution that leads to a different conclusion. 

We adopt DAB-DETR as the basic detection architecture to evaluate our training method, where  the label embedding appended with an indicator is used to replace the decoder embedding part to support label denoising. The difference between our method and other methods is mainly in the training method. In addition to the Hungarian loss, we add a denoising loss as an easier auxiliary task that can accelerate training and boost performance significantly. Chen \etal ~\cite{chen2021pix2seq} augments their sequence with synthetic noise objects, but is totally different from our method. They set the targets of noise objects to the "noise" class (not belonging to any ground-truth classes) so that they can delay the End-of-Sentence (EOS) token and improve the recall. In contrast to their method, we set the target of noised boxes to the original boxes, and the motivation is to bypass bipartite graph matching and directly learn to approximate ground-truth boxes. 

We are pleased to see that many very recent detection models adopt our proposed denoising training to accelerate convergence for detection and segmentation models, such as DINO \cite{zhang2022dino}, Mask DINO \cite{li2022mask}, Group DETR \cite{chen2022group}, and SAM-DETR++ \cite{zhang2022semantic}. DINO \cite{zhang2022dino} further develops our denoising training by feeding hard-negative samples and training the model to reject them. Therefore, the proposed Contrastive Denoising (CDN) further improves the performance. Mask DINO \cite{li2022mask} extends denoising to three image segmentation tasks (instance, panoptic, and semantic) by reconstructing masks from noised boxes. Group DETR \cite{chen2022group} and SAM-DETR+++\cite{zhang2022semantic} also adopt denoising training in their model to achieve better performance. These models demonstrate the effectiveness and generalization capabilities of our methods.
\section{Why Denoising accelerates DETR training?}
\begin{figure}[htbp]
\centering
% \vspace{-0.1cm}
\includegraphics[width=1.07\linewidth]{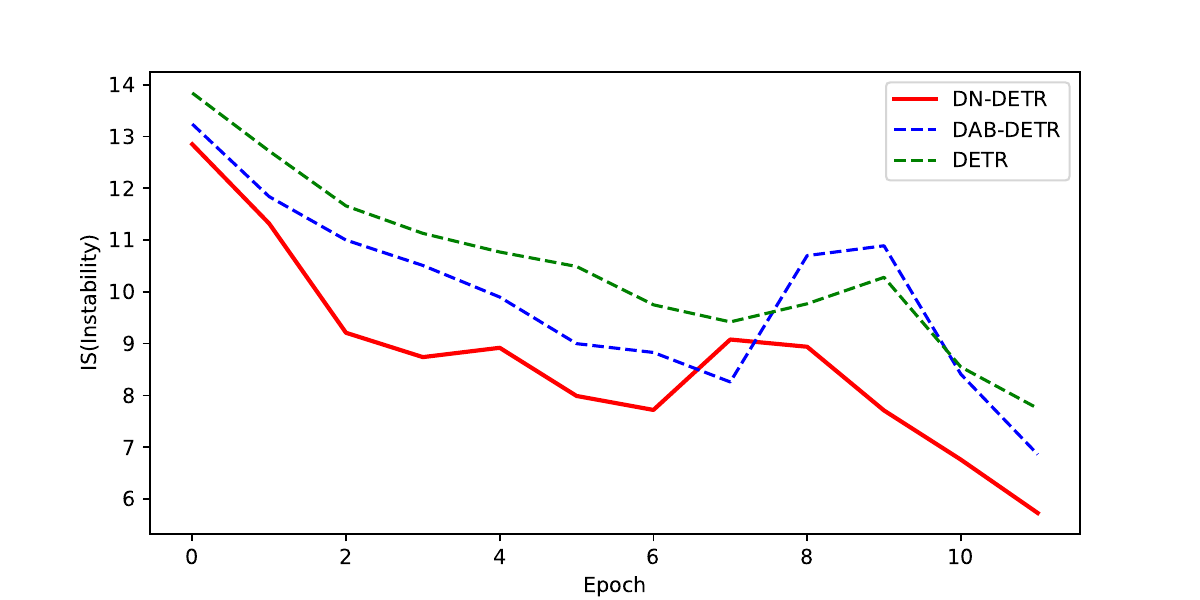}
\centering
% \vspace{-0.7cm}
\caption{The $IS$ of DAB-DETR and DN-DETR during training. For each method, we train $12$ epoch on the same setting. We test the change of the Hungarian matching between each two epochs on the Validation set as the $IS$.}
\label{fig:vib}
\end{figure}
\begin{figure}[htbp]
\centering
% \vspace{-0.1cm}
    \includegraphics[width=0.95\columnwidth]{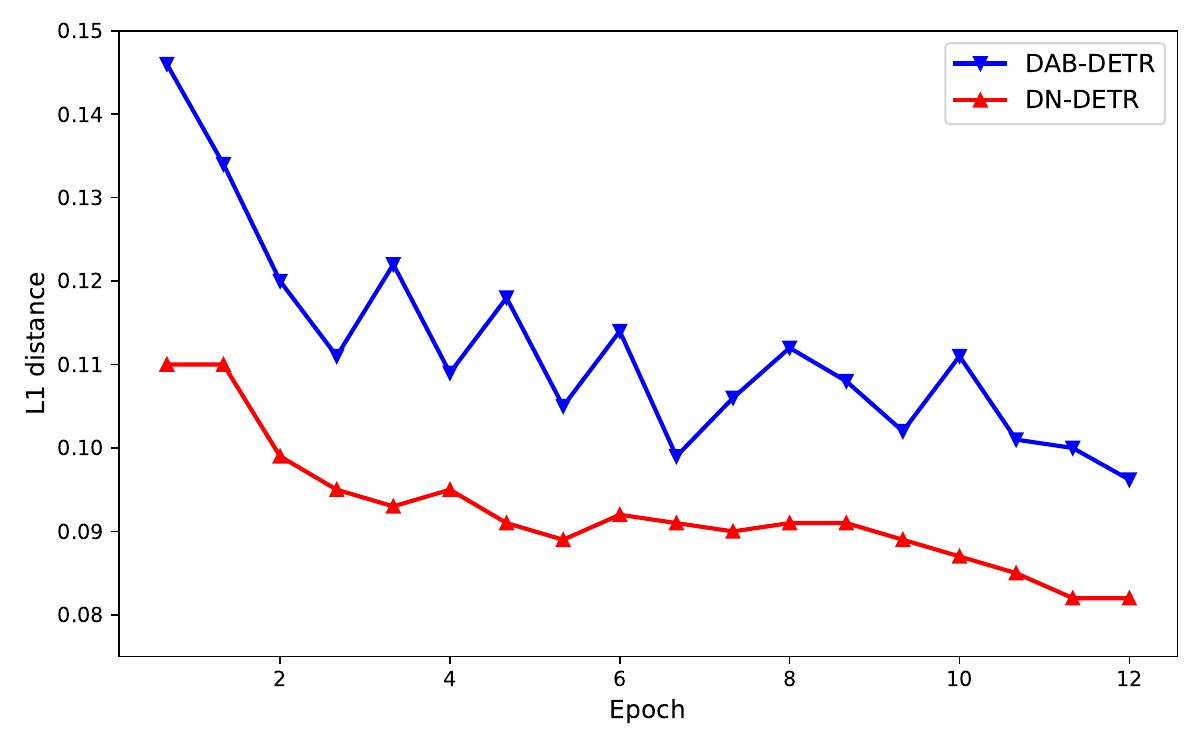}
\centering
% \vspace{-0.7cm}
\caption{A comparison of DAB-DETR and DN-DETR on anchor-target distance.}
\label{fig:vib}
\end{figure}
\begin{figure*}[h]
% \vspace{-0.1cm}
    \includegraphics[width=0.7\columnwidth]{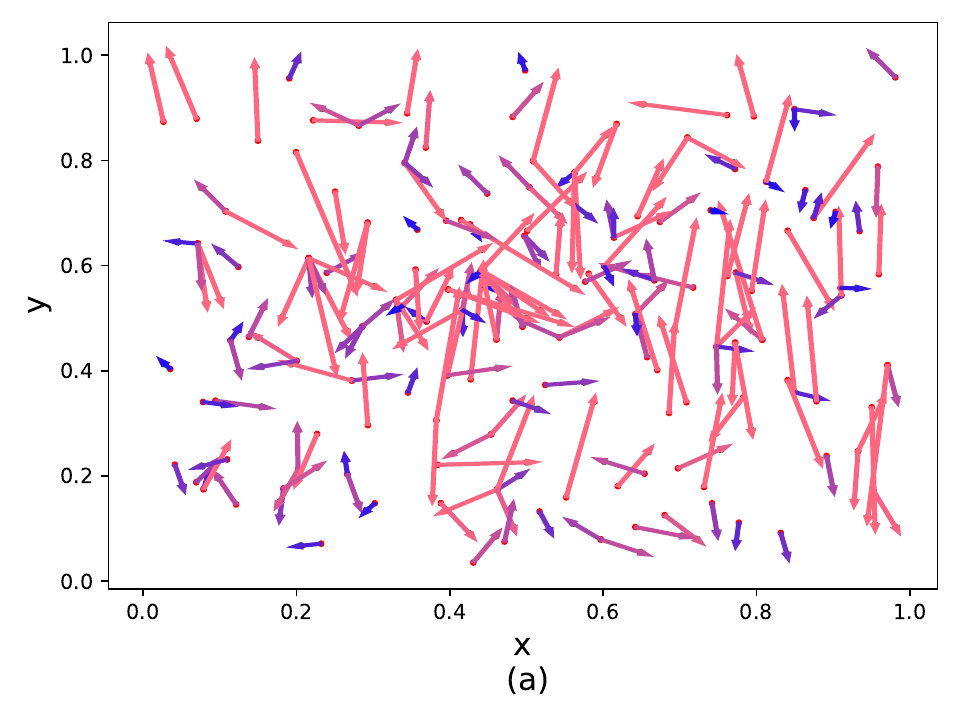}
    \includegraphics[width=0.7\columnwidth]{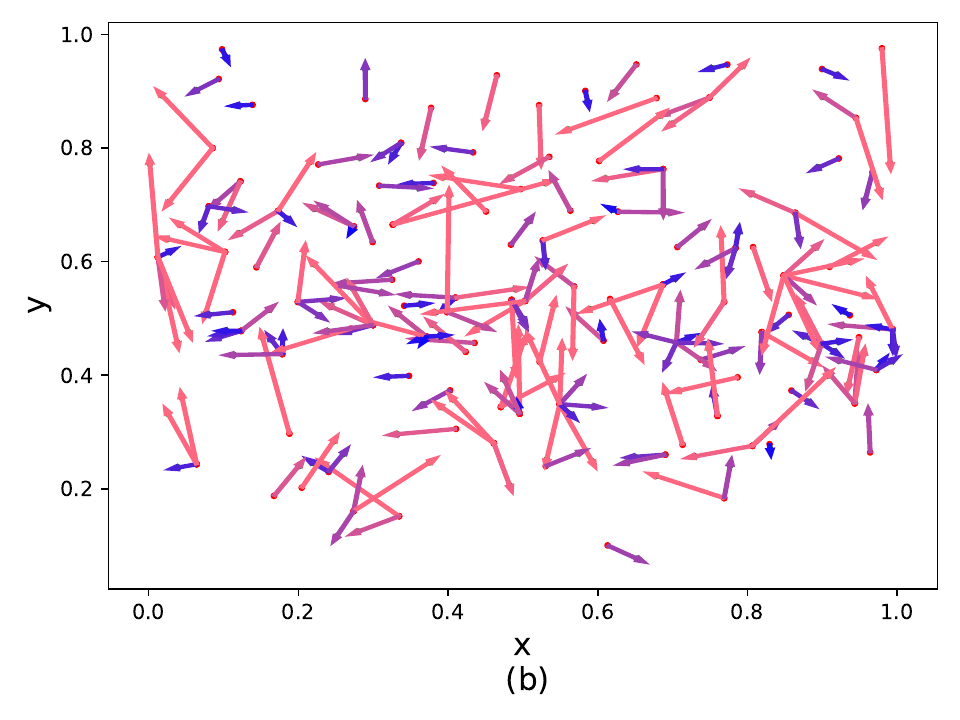}
    \centering
    % \vspace{-0.3cm}
    \caption{\small{ (a)(b)Some examples of anchors and targets for DAB-DETR and DN-DETR, respectively. Each arrow starts from an anchor and points to a target. The color of each arrow shows its \textit{$l_1$} length and cooler colors denote shorter arrows.}}
    \label{fig:distance}
    \vspace{-0.3cm}
\end{figure*}
\subsection{Stablize Hungarian Mathcing}
Hungarian matching is a popular algorithm in graph matching. Given a cost matrix, the algorithm outputs an optimal matching result. DETR is the first algorithm that adopts Hungarian matching in object detection to solve the matching problem between predicted objects and ground-truth objects. DETR turns ground-truth assignment into a dynamic process, which brings in an instability problem due to its discrete bipartite matching and the stochastic training process. There are works \cite{fenoaltea2021stable} showing that Hungarian matching does not result in stable matching since blocking pairs exist. A small change in the cost matrix may cause an enormous change in the matching result, which will further lead to inconsistent optimization goals for decoder queries.

We view the training process of DETR-like models as two stages, learning ``good anchors" and learning relative offsets. Decoder queries are responsible for learning anchors as shown in previous works ~\cite{liu2022dabdetr} and~\cite{zhu2020deformable}. The inconsistent update of anchors can make it difficult to learn relative offsets. Therefore, in our method, we leverage a denoising task as a training shortcut to make relative offset learning easier, as the denoising task bypasses bipartite matching. Since we interpret each decoder query as a $4$-D anchor box, a noised query can be regarded as a ``good anchor" which has a corresponding ground-truth box nearby. The denoising training thus has a clear optimization goal - to predict the original bounding box, which essentially avoids the ambiguity brought by Hungarian matching. 

To quantitatively evaluate the instability of the bipartite matching result, we design a metric as follows. For a training image, we denote the predicted objects from Transformer decoders as $\mathbf{O^i}=\left\{O_0^i, O_1^i, ..., O_{N-1}^i\right\}$ in the $i$-th epoch, where $N$ is the number of predicted objects, and the ground-truth objects as $\mathbf{T}=\left\{T_0, T_1, T_2, ..., T_{M-1}\right\}$ where $M$ is the number of ground-truth objects. After bipartite matching, we compute an index vector $\mathbf{V^i}=\left\{V^i_0, V_1^i, ..., V_{N-1}^i\right\}$ to store the matching result of epoch $i$ as follows.
\begin{equation}
    V^i_n=\left\{\begin{array}{ll}
 m, &\text{if } O^i_n \text{ matches } T_m\\
 -1, &\text{if } O^i_n \text{ matches nothing}
\end{array}\right.
\label{eq: matching}
\end{equation}
We define the instability of epoch $i$ for one training image as the difference between its $V^i$ and $V^{i-1}$, which is calculated as
\begin{equation}
    IS^i=\sum_{j=0}^{N}\mathbbm{1}(V^i_n\neq V^{i-1}_n) 
    \label{eq: instability}
\end{equation}
where $\mathbbm{1}(\cdot)$ is the indicator function. $\mathbbm{1}(x)=1$ if $x$ is true and $0$ otherwise. The instability of epoch $i$ for the whole data set is averaged over the instability numbers for all images. We omit the index for an image for notation simplicity in Eq. \eqref{eq: matching} and Eq. \eqref{eq: instability}.

Fig.~\ref{fig:vib} shows a comparison of $IS$ between our DN-DETR (DeNoising DETR) and DAB-DETR. We conduct this evaluation on the COCO 2017 validation set ~\cite{lin2015microsoft}, which has $7.36$ objects per image on average. So the largest possible $IS$ is $7.36 \times 2=14.72$. Fig. \ref{fig:vib} clearly shows that our method effectively alleviates the instability of matching. 
\subsection{Make Query Search More Locally}
We also show that DN-DETR can help detection by reducing the distance between anchors and the corresponding targets. DETR \cite{carion2020end} shows from the visualization that its positional queries have several operating modes, which makes a query search from a wide region for a predicted box. However, DN-DETR has much smaller mean distances between initial anchors (positional queries) and targets. As shown in Fig.~\ref{fig:distance}(a), we compute the mean $\textit{l}_1$ distance between initial anchors and the matched ground-truth boxes in the last decoder layer for DAB-DETR and our model.  

As denoising training trains the model to reconstruct boxes from the noised ones that are close to the ground truth, the model will search more locally for prediction, which makes each query focus on regions nearby and prevents potential prediction conflicts between queries.   Fig.~\ref{fig:distance}(b) and (c) are some examples of anchors and targets in DAB-DETR and DN-DETR. Each arrow starts from an anchor and ends with its matched ground-truth box. We use color to reflect the length of the arrows. The shortened distances between anchors and targets make the training process easier and therefore converge faster.
\section{DN-DETR}
\begin{figure*}[htbp]
\centering
% \vspace{-0.1cm}
\includegraphics[width=0.65\linewidth]{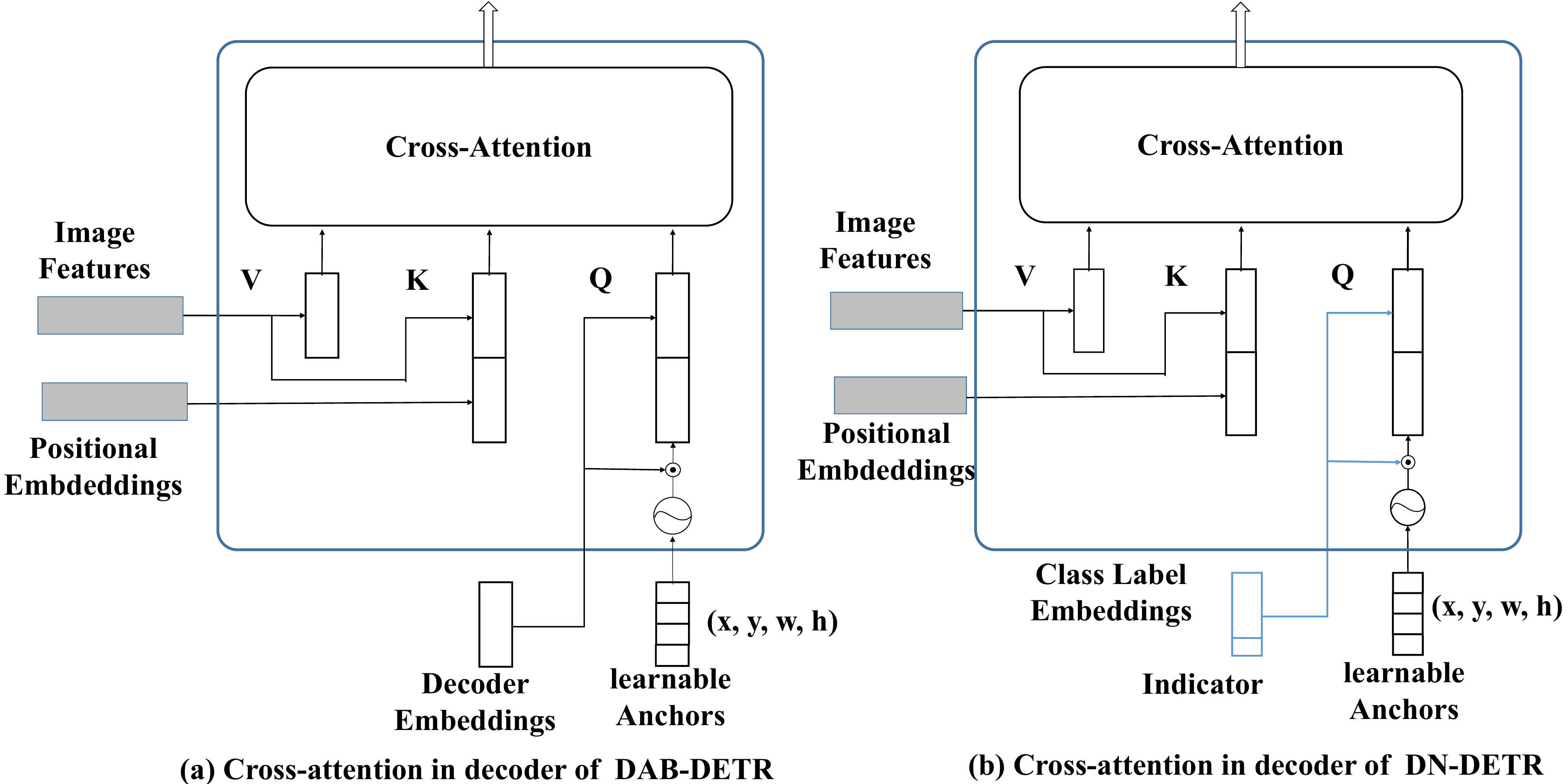}
\centering
% \vspace{-0.7cm}
\caption{Comparison of the cross-attention part DAB-DETR and our DN-DETR (a)DAB-DETR directly uses dynamically updated anchor boxes to provide both a reference query point $(x,y)$ and a reference anchor size $(w,h)$ to improve the cross-attention computation. (b)  DN-DETR specifies the decoder embeddings as label embeddings and adds an indicator to differentiate the denoising task and matching task. }
\label{fig:cross-attention}
\end{figure*}
\begin{figure*}
    \centering
    \includegraphics[width=\linewidth]{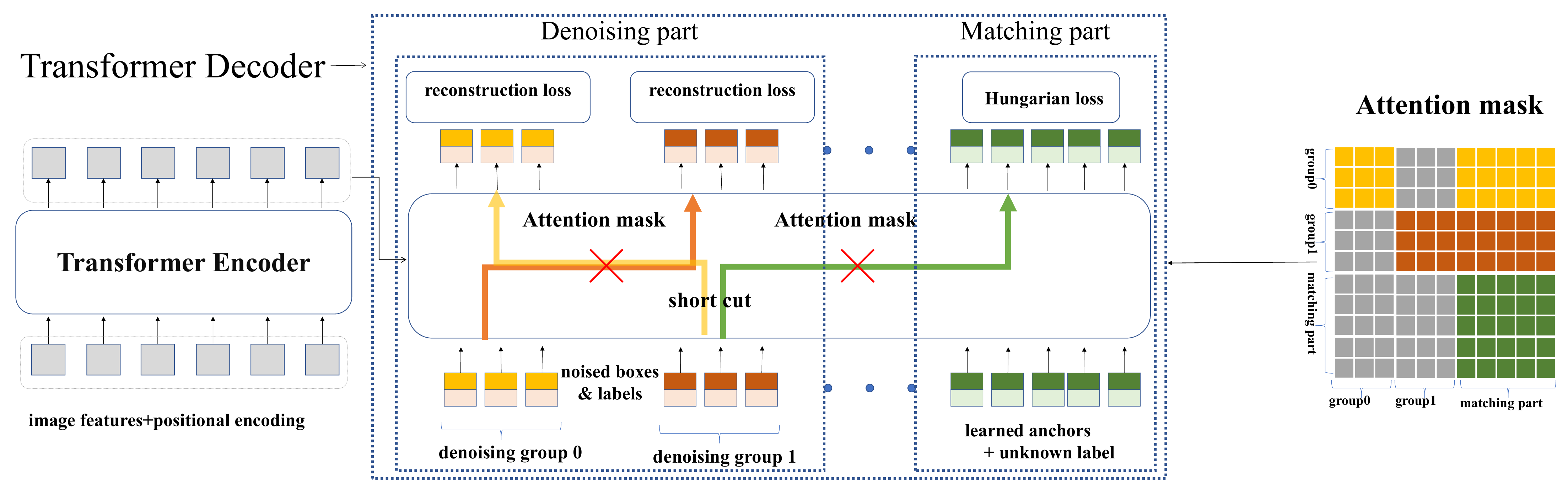}
    % \vspace{-0.2cm}
    \caption{The overview of our training method. There are two parts of queries, namely the denoising part and the matching part. The denoising part contains $\geq 1$ denoising groups. The attention masks from the matching part to the denoising part and among denoising groups are set to $1$ (block) to block information leakage. In the figure, the yellow, brown and green grids in the attention mask represent $0$ (unblock) and grey grids represent $1$ (block). }
    \label{fig:arch}
\end{figure*}

\subsection{Overview}
We base on the architecture of DAB-DETR ~\cite{liu2022dabdetr} to implement our training method. Similar to DAB-DETR, we explicitly formulate the decoder queries as box coordinates. The only difference between our architecture and theirs lies in the decoder embedding, which is specified as class label embedding to support label denoising. Our main contribution is the training method as shown in Fig. \ref{fig:arch}.

Similar to DETR, our architecture contains a Transformer encoder and a Transformer decoder. On the encoder side, the image features are extracted with a CNN backbone and then fed into the Transformer encoder with positional encodings to attain refined image features. On the decoder side, queries are fed into the decoder to search for objects through cross-attention. 

We denote decoder queries as $\mathbf{q}=\left\{q_0, q_1, ..., q_{N-1}\right\}$ and the output of the Transformer decoder as $\mathbf{o}=\left\{o_0, o_1, ..., o_{N-1}\right\}$. We also use $F$ and $A$ to denote the refined image features after the Transformer encoder, and the attention mask derived based on the denoising task design. We can formulate our method as follows.
\begin{equation}
    \mathbf{o}=D(\mathbf{q},F|A)
    % \vspace{-0.2cm}
\end{equation}
where $D$ denotes the Transformer decoder. 

There are two parts to decoder queries. One is the matching part. The inputs of this part are learnable anchors, which are treated in the same way as in DETR. That is, the matching part adopts bipartite graph matching and learns to approximate the ground-truth box-label pairs with matched decoder outputs. The other is the denoising part. The inputs of this part are noised ground-truth (GT) box-label pairs which are called \textbf{GT objects} in the rest of the paper. The outputs of the denoising part aim to reconstruct GT objects.

In the following, we abuse the notations to denote the denoising part as  $\mathbf{q}=\left\{q_0, q_1, ..., q_{K-1}\right\}$ and the matching part as $\mathbf{Q}=\left\{ Q_0, Q_1, ..., Q_{L-1}\right\}$. So the formulation of our method becomes
\begin{equation}
    \mathbf{o}=D(\mathbf{q},\mathbf{Q},F|A)
    % \vspace{-0.2cm}
\end{equation}
To increase the denoising efficiency, we propose to use multiple versions of noised GT objects in the denoising part.
Furthermore, we utilize an attention mask to prevent information leakage from the denoising part to the matching part and among different noised versions of the same GT object.
\subsection{Intro to DAB-DETR}
Many recent works associate DETR queries with different positional information. DAB-DETR follows this analysis and explicitly formulates each query as 4D anchor coordinates. As shown in Fig. \ref{fig:cross-attention}(a), a query is specified as a tuple $(x,y,w,h)$, where $x,y$ are the center coordinates and $w,h$ are the corresponding width and height of each box. In addition, the anchor coordinates are dynamically updated layer by layer. The output of each decoder layer contains a tuple $(\Delta x,\Delta y,\Delta w,\Delta h)$ and the anchor is updated to $(x + \Delta x, y+\Delta y,w+\Delta w,h+\Delta h)$.
\par
Note that our proposed method is mainly a training method that can be integrated into any DETR-like model. To test on DAB-DETR, we only add minimal modifications: specifying the decoder embedding as label embedding, as shown in Fig. \ref{fig:cross-attention}(b). 
\subsection{Denoising}
\label{denoising}
For each image, we collect all GT objects and add random noises to both their bounding boxes and class labels. To maximize the utility of denoising learning, we use multiple noised versions for each GT object. 

We consider adding noise to boxes in two ways: center shifting and box scaling. We define $\lambda_1$ and $\lambda_2$ as the noise scale of these $2$ noises. 1) \textbf{center shifting:} we add a random noise $(\Delta x, \Delta y)$, to the box center and make sure that $|\Delta x|<\frac{\lambda_1 w}{2}$ and $|\Delta y|<\frac{\lambda_1 h}{2}$, where $\lambda_1\in(0,1)$ so that the center of the noised box will still lie inside the original bounding box. 2) \textbf{box scaling:} we set a hyper-parameter $\lambda_2 \in (0, 1)$. The width and height of the box are randomly sampled in $\left[(1-\lambda_2)w, (1+\lambda_2)w\right]$ and $\left[(1-\lambda_2)h, (1+\lambda_2)h\right]$, respectively. 

For label noising, we adopt label flipping, which means we randomly flip some GT labels to other labels. Label flipping forces the model to predict the GT labels according to the noised boxes to better capture the label-box relationship. We have a hyper-parameter $\gamma$ to control the ratio of labels to flip. The reconstruction losses are $l_1$ loss and GIOU loss for boxes and focal loss ~\cite{lin2018focal} for class labels as in DAB-DETR. We use a function $\delta(\cdot)$ to denote the noised GT objects. Therefore, each query in the denoising part can be represented as $q_k=\delta(t_m)$ where $t_m$ is $m$-th GT object.

Notice that denoising is only considered in training, during inference the denoising part is removed, leaving only the matching part.

\subsection{Attention Mask}
Attention mask is a component of great importance in our model.
Without an attention mask, the denoising training will compromise the performance instead of improving it as shown in Table \ref{tab:ablation}. 

To introduce an attention mask, we need first to divide the noised GT objects into groups. Each group is a noised version of all GT objects. The denoising part becomes
\begin{equation}
% \vspace{-0.2cm}
\mathbf{q}=\left\{\mathbf{g_0}, \mathbf{g_1}, ..., \mathbf{g_{P-1}}\right\}
\end{equation}
where $\mathbf{g_p}$ is defined as the $p$-th \textbf{denoising group}. Each denoising group contains $M$ queries where $M$ is the number of GT objects in the image. So we have
\begin{equation}
% \vspace{-0.2cm}
\mathbf{g_p}=\left\{q^p_0, q^p_1, ..., q^p_{M-1} \right\}
\end{equation}
where $q^p_m=\delta(t_m)$.

The purpose of the attention mask is to prevent information leakage. There are two types of potential information leakage. One is that the matching part may see the noised GT objects and easily predict GT objects. The other is that one noised version of a GT object may see another version. Therefore, our attention mask is to make sure the matching part cannot see the denoising part and the denoising groups cannot see each other as shown in Fig. \ref{fig:arch}. 

We use $\mathbf{A}=\left[\mathbf{a}_{i j}\right]_{W \times W}$ to denote the attention mask where $W=P\times M+N$. $P$ and $M$ are the number of groups and GT objects. $N$ is the number of queries in the matching part. We let the first $P\times M$ rows and columns represent the denoising part and the latter represents the matching part. $a_{ij}=1$ means the $i$-th query cannot see the $j$-th query and $a_{ij}=0$ otherwise. We devise the attention mask as follows
\begin{equation}
a_{ij}=\left\{\begin{array}{ll}
1, & \text { if } j<P\times M \text{ and } \lfloor\frac{i}{M}\rfloor \neq \lfloor\frac{j}{M}\rfloor;\\
1, & \text { if } j<P\times M \text{ and } i\geq P\times M;\\
0, & \text{otherwise.}
\end{array}\right.
% \vspace{-0.2cm}
\end{equation}
% Note that we allow the denoising part to see the matching part, since the queries of matching part are learned queries that contain no information of the ground truth objects. 
Note that whether the denoising part can see the matching part or not will not influence the performance, since the queries of the matching part are learned queries that contain no information about the GT objects. 

The extra computation introduced by multiple denoising groups is negligible---when $5$ denoising groups are introduced, GFLOPs for training are only increased from $94.4$ to $94.6$ for DAB-DETR with a ResNet-50 backbone, and there is no computation overhead for testing.

\subsection{Label Embedding}
The decoder embedding is specified as label embedding in our model to support both box denoising and label denoising. Except for the $80$ classes in COCO 2017 ~\cite{lin2015microsoft}, we also consider an unknown class embedding that is used in the matching part to be semantically consistent with the denoising part. We also append an indicator to label embedding. The indicator is $1$ if a query belongs to the denoising part and $0$ otherwise. 

\subsection{Compatibility with Deformable Attention Design}
\textbf{DN-Deformable-DETR:} To show the effectiveness of denoising training applied in other attention designs, we also integrate denoising training into Deformable DETR as {DN-Deformable-DETR}. We follow the same setting as Deformable DETR but specify its query into $4$D boxes as in DAB-DETR to better use denoising training. Note that this is our original deformable model in the conference version, in which we only add deformable attention to Transformer encoders.

When comparing in the standard 50 epoch setting, to eliminate any misleading information that the performance improvement of DN-Deformable-DETR may result from the explicit query formulation of anchor boxes, we also implement a strong baseline DAB-Defromable-DETR for comparison. It formulates the queries of Deformable DETR as anchor boxes without using denoising training, while all the other settings are the same.\\
\textbf{DN-Deformable-DETR++:} We further incorporate the deformable attention in our decoder and optimize our model to build DN-Deformable-DETR++, which converges much faster and improves the final results. We also follow DAB-Defromable-DETR to build a strong baseline DAB-Defromable-DETR++ to show our performance improvement in the ablations.
\begin{table*}[t]
    \centering
        \caption{Results for our DN-DETR and other detection models under the same setting. All DETR-like models except DETR use $300$ queries, while DETR uses $100$. 
    % The models with superscript $^{*}$ use $3$ pattern embeddings as in Anchor DETR~\cite{wang2021anchor}.
    }
   \resizebox{1\textwidth}{!}{%
    \begin{tabular}{lccccccccccc}
        \toprule
        Model & \#epochs & AP & AP$_{50}$ & AP$_{75}$ & AP$_{S}$ & AP$_{M}$ & AP$_{L}$ & GFLOPs & Params \\
        \midrule
        DETR-R$50$~\cite{carion2020end}                & $500$ & $42.0$ & $62.4$ & $44.2$ & $20.5$ & $45.8$ & $61.1$ & $86$ & $41$M \\
        Faster RCNN-FPN-R$50$~\cite{ren2015faster}   & $108$ & $42.0$  & $62.1$ & $45.5$ & $26.6$ & $45.5$ & $53.4$ & $180$ & $42$M \\
        Anchor DETR-R$50$~\cite{wang2021anchor}         & $50$ & $42.1$ & $63.1$ &  $44.9$ &  $22.3$ &  $46.2$  & $60.0$ & $-$ & $39$M \\
        Conditional DETR-R$50$~\cite{meng2021conditional}    & $50$ & $40.9$ & $61.8$ & $43.3$ & $20.8$ & $44.6$ & $59.2$ & $90$ & $44$M \\
        DAB-DETR-R$50$~\cite{liu2022dabdetr}    & $50$ & $42.2$ & $63.1$ & $44.7$ & $21.5$ & $45.7$ & $60.3$ & $94$ & $44$M \\
        DN-DETR-R$50$& $50$ & $\textbf{44.1(+1.9)}$ & $64.4$ & $46.7$ & $22.9$ & $48.0$ & $63.4$ & $94$ & $44$M \\
        \hline
        DETR-R$101$~\cite{carion2020end}                  & $500$ & $43.5$ & $63.8$ & $46.4$ & $21.9$ & $48.0$ & $61.8$ & $152$ & $60$M \\
        Faster RCNN-FPN-R$101$~\cite{ren2015faster}  & $108$ & $44.0$ & $63.9$ & $47.8$ & $27.2$ & $48.1$ & $56.0$ & $246$ & $60$M \\
        Anchor DETR-R$101$~\cite{wang2021anchor}           & $50$ & $43.5$ &  $64.3$ &  $46.6$ &  $23.2$  & $47.7$  & $61.4$ & $-$ & $58$M \\
        Conditional DETR-R$101$~\cite{meng2021conditional}   & $50$ & $42.8$ &  $63.7$ &  $46.0$ &  $21.7$ &  $46.6$ &  $60.9$  & $156$ & $63$M\\
        DAB-DETR-R$101$~\cite{liu2022dabdetr}     & $50$ & {${43.5}$} & $63.9$ & $46.6$ & $23.6$ & $47.3$ & $61.5$ & $174$ & $63$M \\
        DN-DETR-R$101$    & $50$ & {$\textbf{45.2(+1.7)}$} & $65.5$ & $48.3$ & $24.1$ & $49.1$ & $65.1$ & $174$ & $63$M \\
        \hline
        DETR-DC5-R$50$~\cite{carion2020end}               & $500$ & $43.3$ & $63.1$ & $45.9$ & $22.5$ & $47.3$ & $61.1$ & $187$ & $41$M \\
        Anchor DETR-DC5-R$50$~\cite{wang2021anchor}           & $50$ & $44.2$ & $64.7$ & $47.5$ & $24.7$ & $48.2$ & $60.6$ & $151$ & $39$M \\
        Conditional DETR-DC5-R$50$~\cite{meng2021conditional}  & $50$  &  $43.8$ & $64.4$ &  $46.7$ &  $24.0$ & $47.6$ &  $60.7$ & $195$ & $44$M \\
        DAB-DETR-DC5-R$50$~\cite{liu2022dabdetr}    & $50$ & {${44.5}$} & $65.1$ & $47.7$ & $25.3$ & $48.2$ & $62.3$ & $202$ & $44$M \\
        DN-DETR-DC5-R$50$    & $50$ & $\textbf{46.3(+1.8)}$ & $66.4$ & $49.7$ & $26.7$ & $50.0$ & $64.3$ & $202$ & $44$M \\
        \hline 
        DETR-DC5-R$101$~\cite{carion2020end}   & $500$ & $44.9$ & $64.7$ & $47.7$ & $23.7$ & $49.5$ & $62.3$ & $253$ & $60$M \\
        Anchor DETR-R$101$~\cite{wang2021anchor}           & $50$ & $45.1$ & $65.7$ & $48.8$ & $25.8$ & $49.4$ & $61.6$ & $-$ & $58$M \\
        Conditional DETR-DC5-R$101$~\cite{meng2021conditional}  & $50$  &   $45.0$ & $65.5$ & $48.4$ & $26.1$ & $48.9$ & $62.8$ & $262$ & $63$M \\
        DAB-DETR-DC5-R$101$~\cite{liu2022dabdetr}  & $50$ & {${45.8}$} & $65.9$ & $49.3$ & $27.0$ & $49.8$ & $63.8$ & $282$ & $63$M \\
        DN-DETR-DC5-R$101$    & $50$ & $\textbf{47.3(+1.5)}$ & $67.5$ & $50.8$ & $28.6$ & $51.5$ & $65.0$ & $282$ & $63$M \\
        \bottomrule
    \end{tabular}}
    
    \centering

    \label{tab:absoulute_results}
    % \vspace{-.3cm}
\end{table*}
\subsection{Introducing DN to Other DETR-like models with different anchor formulations}
In the aforementioned sections, we build DN-DETR upon DAB-DETR \cite{liu2022dabdetr} with explicit 4D anchor box formulation.
As shown in Fig. \ref{fig:arch}, denoising is only a training method and can be plugged into other detection models to accelerate training. In this section, we will extend denoising training to other DETR-like models. 
\subsubsection{Introducing DN to Anchor DETR with 2D Anchors}
We first demonstrate its effectiveness by adding it to Anchor DETR \cite{wang2021anchor}, which formulates positional queries as 2D anchor points.
For DN-Anchor-DETR, though it can be easily modified to 4D anchors to achieve better results, we strictly follow Anchor DETR to add noise only to 2D anchors. A 2D anchor corresponds to the center point of a box. Hence we only use center shifting noise (described in Sec. \ref{denoising}). In this way, we plug in the denoising training task for anchor points without introducing other modifications.
\subsubsection{Introducing DN to Vanilla DETR without Explicit Anchors}
Vanilla DETR \cite{carion2020end} differs from DAB-DETR in that its positional queries are
high dimensional vectors without explicit meanings.
For DN-Vanilla-DETR, we can simply use linear box embedding to embed noised boxes into the same dimension as DETR queries. The content query part is the same as DAB-DETR, and we use label embedding to embed labels into content queries. After obtaining content and position queries, following Vanilla DETR, we can add the label embedding and box embedding together as DETR queries.
\subsection{Introducing DN to Faster R-CNN for Traditional Detectors}
Apart from accelerating DETR-like models, denoising training can also be used to accelerate traditional CNN detectors. We take Faster R-CNN \cite{ren2015faster} as an example and add denoising training to it. The detection head of Faster R-CNN works in a similar way as the decoder of DETR-based models, where the major differences lie in  1) feature extraction: Faster R-CNN uses RoI pooling  while DETR uses cross attention to extract features.  and 2) label-assignment scheme: Faster R-CNN adopts a one-to-many label assignment (one GT object can be matched with multiple predicted objects), while DETR adopts a one-to-one label assignment (one GT object can only be matched with one predicted object). As the denoising part trains in parallel with the original matching part in detection models and is irrelevant to feature extraction schemes, denoising training can be easily applied to these traditional detectors.  

Fundamentally, the idea of denoising training in DETR is to bypass the unstable label assignment and directly learn bounding box regression. Though Faster R-CNN does not have bipartite matching, it also has label assignment controlled by the IoU threshold. Therefore, denoising training can also serve as a shortcut to help learn bounding box regression without label assignment in traditional models. Therefore, we add noised boxes to the detection head of Faster R-CNN in parallel with the original boxes from the RPN. These noised boxes will directly regress the GT to improve training. Note that as Faster R-CNN does not have an initial content part, we only use box denoising training.
\subsection{Introducing DN to Mask2Former for Segmentation Models}
We also show the feasibility of adding denoising training to segmentation models such as Mask2Former \cite{cheng2022masked}. Mask2Former adopts a DETR-like architecture and proposes masked attention to extract features for segmentation tasks. More specifically, each decoder layer predicts segmentation masks, which are passed to the subsequent decoder layer as the attention mask to pool features. Therefore, following the idea of denoising training in detection models, we can add noise to the GT masks and feed them to the decoder as the attention mask. The training objective of these noised masks is to directly predict the GT mask, which bypasses the bipartite match and serves as a shortcut to directly learn mask refinement.

To verify the effectiveness of denoising training on masks, we build a simple baseline by adding simple
shifting noise to the mask. Without changing the shape or size of the mask, we shift the whole GT mask on the x-axis and y-axis by a random value, which is the same as the center shifting noise as described in Sec. \ref{denoising}. This simple baseline already demonstrates the effectiveness of denoising training.
\section{Experiment}
\subsection{Setup}
\label{Sec:Setup}
\noindent\textbf{Dataset:} We show the effectiveness of DN-DETR on the challenging MS-COCO 2017 ~\cite{lin2015microsoft} Detection task. MS-COCO is composed of 160K images with 80 categories. These images are divided into \texttt{train2017} with 118K images, \texttt{val2017} with 5K images, and \texttt{test2017} with 41K images. In all our experiments, we train the models on \texttt{train2017} and test on \texttt{val2017}.
Following the common practice, we report the standard mean average precision (AP) result on the COCO validation dataset under different IoU thresholds and object scales.
\\
\textbf{Implementation Details:} 
We test the effectiveness of the denoising training on DAB-DETR, which is composed of a CNN backbone, multiple Transformer encoder layers, and decoder layers. We also show that denoising training can be plugged into other DETR-like models to boost performance. For example, our DN-Deformable-DETR is built upon Deformable DETR in a multi-scale setting.
\par

\begin{table*}[t]
    \centering
    %     \footnotesize
    %         \renewcommand{\arraystretch}{1.6}
        \caption{Results for our DN-DETR and other detection models on the 1x setting. Superscript $^\dag$ indicates that we check with the authors of Dynamic DETR through private communication, their encoder deign makes their single-scale and multi-scale results almost identical.
    }
    \resizebox{1\textwidth}{!}
    {
    \begin{tabular}{lcccccccccc}
        \toprule
        Model & MultiScale & \#epochs & AP & AP$_{50}$ & AP$_{75}$ & AP$_{S}$ & AP$_{M}$ & AP$_{L}$ & GFLOPs & Params \\
        \midrule
        % Faster R$50$-DC5 1x&& $12$ & $37.3$ & \\
        Faster R-CNN-FPN-R$50$ 1x \cite{ren2015faster}&$\checkmark$& $12$ & $37.9$ & $58.8$ & $41.1$ & $22.4$ & $41.1$ & $49.1$ &$180$ & $40$M\\
        % Faster R$50$-FPN 1x& $12$ & $37.9$ & $--$ & $--$ & $--$ & $--$ & $--$ &$--$ & $--$M\\
        DETR-R$50^{}$ 1x \cite{carion2020end}&& $12$ & $15.5$ & $29.4$ & $14.5$ & $4.3$ & $15.1$ & $26.7$ &$86$ & $41$M\\
        DAB-DETR-DC5-R$50$ \cite{liu2022dabdetr}&& $12$ & $38.0$ & $60.3$ & $39.8$ & $19.2$ & $40.9$ & $55.4$ &$216$ & $44$M\\
        DN-DETR-DC5-R$50$&& $12$ & $\textbf{41.7(+3.7)}$ & $61.4$ & $44.1$ & $21.2$ & $45.0$ & $60.2$ &$216$ & $44$M\\
        \hline
        % BorderDet-R50 \cite{qiu2020borderdet}&&12&41.4& 59.4& 44.5& 23.6& 45.1& 54.6&-&-
        % \\
        Deformable DETR-R$50^{}$ 1x \cite{zhu2020deformable}&$\checkmark$& $12$ & $37.2$ & $55.5$ & $40.5$ & $21.1$ & $40.7$ & $50.5$ &$173$ & $40$M\\
        % Dynamic DETR-R$50^{\dag}$ \\(no dynamic encoder)
        \begin{tabular}[c]{@{}c@{}}Dynamic DETR-R$50^{\dag}$ 1x \\ w/o dynamic encoder\end{tabular}
        &$\checkmark$& $12$ & $40.2$ & $58.6$ & $43.4$ & $--$ & $-$ & $-$ &$-$ & $-$\\
        Dynamic DETR-R$50^{\dag}$ 1x \cite{dai2021dynamic}&$\checkmark$& $12$ & $42.9$ & $61.0$ & $46.3$ & $24.6$ & $44.9$ & $54.4$ &$-$ & $-$\\
        % DN-DETR-DC5-R$50$& $12$ & $\textbf{40.7}$ & $--$ & $--$ & $--$ & $--$ & $--$ &$--$ & $--$M\\
        DN-Deformable-DETR-R$50$ &$\checkmark$& $12$ & $\textbf{43.4}$ & $61.9$ & $47.2$ & $24.8$ & $46.8$ & $59.4$ &$195$ & $48$M\\
        DN-Deformable-DETR-R$50$++ &$\checkmark$& $12$ & $\textbf{46.0}$ & $63.8$ & $49.9$ & $27.7$ & $49.1$ & $62.3$ &$-$ & $47$M\\
        \hline
        \hline
        % \hline
        DAB-DETR-DC5-R$101$ \cite{liu2022dabdetr}&& $12$ & ${40.3}$ & $62.6$ & $42.7$ & $22.2$ & $44.0$ & $57.3$ &$282$ & $63$M\\
        DN-DETR-DC5-R$101$&& $12$ & $\textbf{42.8(+2.5)}$ & $62.9$ & $45.7$ & $23.3$ & $46.6$ & $61.3$ &$282$ & $63$M\\
        \hline

        Faster R$101$ FPN \cite{ren2015faster}&$\checkmark$ & $108$ & ${44.0}$ & $63.9$ & $47.8$ & $ 27.2$ & $48.1$ & $56.0$ &$246$ & $60$M \\
        
        DN-Deformable-DETR-R$101$&$\checkmark$& $12$ & ${\textbf{44.1}}$ & $62.8$ & $47.9$ & $26.0$ & $47.8$ & $61.3$ &$275$ & $67$M \\
        % \hline
        % \hline
        % % \hline
        % Anchor-DETR-DC5-R$50$ \cite{wang2021anchor}&& $12$ & ${38.2}$ & $62.6$ & $42.7$ & $22.2$ & $44.0$ & $57.3$ &$-$ & $63$M\\
        % DN-Anchor-DETR-DC5-R$50$&& $12$ & $\textbf{39.4(+1.2)}$ & $59.1$ & $41.8$ & $19.6$ & $43.4$ & $56.0$ &$-$ & $37$M\\
        % \hline
        % \hline
        % % \hline
        % Vanilla-DETR-R$50$ \cite{carion2020end}&&$500$ & $42.0$ & $62.4$ & $44.2$ & $20.5$ & $45.8$ & $61.1$ & $86$ & $41$M\\
        % DN-Vanilla-DETR-R$50$&& $250$ & $\textbf{42.2(+0.2)}$ & $59.1$ & $41.8$ & $19.6$ & $43.4$ & $56.0$ &$86$ & $37$M\\
        % \hline
        % \hline
        % % \hline
        % Faster R-CNN-FPN-R$50$ \cite{wang2021anchor}&$\checkmark$& $12$ & ${37.9}$ & $58.8$ & $41.1$ & $22.4$ & $41.1$ & $49.1$ &$180$ & $40$M\\
        % DN-Faster R-CNN-FPN-R$50$&$\checkmark$& $12$ & $\textbf{38.4(+0.5)}$ & $59.1$ & $41.5$ & $22.7$ & $41.6$ & $50.4$ &$180$ & $40$M\\
        % \hline
        % \hline
        % % \hline
        % SAM-DETR++-R$50$ \cite{zhang2022semantic}&$\checkmark$& $12$ & ${43.2}$ & $61.5$ & $46.5$ & $25.5$ & $46.5$ & $58.6$ &$203$ & $55$M\\
        % DN-SAM-DETR++-R$50^*$ \cite{zhang2022semantic}&$\checkmark$& $12$ & $\textbf{44.8(+1.6)}$ & $62.6$ & $47.9$ & $26.7$ & $48.2$ & $60.9$ &$203$ & $55$M\\
        % \hline
        % \hline
        % % \hline
        % DINO-R$50$ w/o DN \cite{zhang2022semantic}&$\checkmark$& $12$ & $45.1$ & $46.0$ & $46.5$ & $25.5$ & $46.5$ & $58.6$ &$203$ & $55$M\\
        % DINO-R$50$ w/ DN $^*$ \cite{zhang2022semantic}&$\checkmark$& $12$ & $\textbf{44.8(+1.6)}$ & $62.6$ & $47.9$ & $26.7$ & $48.2$ & $60.9$ &$203$ & $55$M\\
        \bottomrule
    \end{tabular}
    }
    
    \centering

    \label{tab:1x}
    % \vspace{-.3cm}
\end{table*}
\definecolor{Gray}{gray}{0.9}
\begin{table*}[t]
    \centering
    %     \footnotesize
    %         \renewcommand{\arraystretch}{1.6}
        \caption{Extending denoisng training to other detection and segmentation models. Superscript $^*$ means this result is from the ablation experiments of the original paper that uses our denoising training.
    }
    \resizebox{1\textwidth}{!}
    {
    \begin{tabular}{lcccccccccc}
        \toprule
        Model & MultiScale & \#epochs & AP & AP$_{50}$ & AP$_{75}$ & AP$_{S}$ & AP$_{M}$ & AP$_{L}$ & GFLOPs & Params \\
        \toprule[1.2pt]
        \rowcolor{Gray}\multicolumn{2}{c}{\textbf{Extending DN to other detection models}} &&&&&&&&&\\

        Anchor-DETR-DC5-R$50$ \cite{wang2021anchor}&& $12$ & ${38.2}$ & $58.6$ & $40.6$ & $20.3$ & $41.9$ & $53.1$ &$-$ & $37$M\\
        DN-Anchor-DETR-DC5-R$50$&& $12$ & $\textbf{39.4(+1.2)}$ & $59.1$ & $41.8$ & $19.6$ & $43.4$ & $56.0$ &$-$ & $37$M\\
        \hline
        \hline
        Group-DAB-DETR-DC5-R$50$ \cite{chen2022group}&& $12$ & ${41.9}$ & $-$ & $-$ & $23.3$ & $45.6$ & $58.4$ &$-$ & $-$M\\
        DN-Group-DAB-DETR-DC5-R$50^*$ \cite{chen2022group}&& $12$ & $\textbf{44.5(+2.6)}$ & $-$ & $-$ & $25.9$ & $48.2$ & $62.2$ &$-$ & $-$M\\
        \hline
        \hline
        % \hline
        % \hline
        Faster R-CNN-FPN-R$50$ \cite{wang2021anchor}&$\checkmark$& $12$ & ${37.9}$ & $58.8$ & $41.1$ & $22.4$ & $41.1$ & $49.1$ &$180$ & $40$M\\
        DN-Faster R-CNN-FPN-R$50$&$\checkmark$& $12$ & $\textbf{38.4(+0.5)}$ & $59.1$ & $41.5$ & $22.7$ & $41.6$ & $50.4$ &$180$ & $40$M\\
        \hline
        \hline
        % \hline
        SAM-DETR++-R$50$ \cite{zhang2022semantic}&$\checkmark$& $12$ & ${43.2}$ & $61.5$ & $46.5$ & $25.5$ & $46.5$ & $58.6$ &$203$ & $55$M\\
        DN-SAM-DETR++-R$50^*$ \cite{zhang2022semantic}&$\checkmark$& $12$ & $\textbf{44.8(+1.6)}$ & $62.6$ & $47.9$ & $26.7$ & $48.2$ & $60.9$ &$203$ & $55$M\\
        \hline
        \hline
        % \hline
        DINO-R$50$ w/o DN \cite{zhang2022dino}&$\checkmark$& $12$ & $46.0$ & $64.0$ & $49.9$ & $29.3$ & $49.2$ & $60.5$ &$279$ & $47$M\\
        DINO-R$50$ w/ DN$^*$ \cite{zhang2022dino}&$\checkmark$& $12$ & $\textbf{47.4(+1.4)}$ & $64.6$ & $51.3$ & $30.0$ & $50.7$ & $61.8$ &$279$ & $47$M\\
        \hline
        \hline
        Vanilla-DETR-R$50$ \cite{carion2020end}&&$300$ & $40.6$ & $61.6$ & $-$ & $19.9$ & $44.3$ & $60.2$ & $86$ & $41$M\\
        DN-Vanilla-DETR-R$50$&& $300$ & $\textbf{42.6(+2.0)}$ & $62.3$ & $44.9$ & $21.6$ & $46.1$ & $61.4$ &$86$ & $37$M\\
        \toprule[1.2pt]
        \rowcolor{Gray}\multicolumn{2}{c}{\textbf{Extending DN to segmentation models}} &&&&&&&&&\\

        Mask DINO-R$50$ w/o mask DN \cite{li2022mask}&$\checkmark$& $12$ & $40.7$ & $62.8$ & $43.7$ & $21.0$ & $43.4$ & $60.6$ &$234$ & $50$M\\
        Mask DINO-R$50$ w/ mask DN $^*$ \cite{li2022mask}&$\checkmark$& $12$ & $\textbf{41.4(+0.7)}$ & $62.9$ & $44.6$ & $21.1$ & $44.2$ & $61.4$ &$234$ & $50$M\\
        \hline
        \hline
        Mask2Former-R$50$ \cite{cheng2022masked}&$\checkmark$& $12$ &  $38.7$ & $59.8$ &$41.2$& $18.2$ & $41.5$ & $59.8$ &$226$ & $44$M\\
        DN-Mask2Former-R$50$ &$\checkmark$& $12$ & $\textbf{39.7(+1.0)}$ & $60.8$ & $42.3$ & $19.1$ & $42.7$ & $61.2$ &$226$ & $44$M\\

        \bottomrule
    \end{tabular}
    }
    
    \centering

    \label{tab:dn_series}
    % \vspace{-.3cm}
\end{table*}
\begin{table*}[t]
    \centering
    %     \footnotesize
    %         \renewcommand{\arraystretch}{1.6}
    % \resizebox{\columnwidth}{!}{%
        \caption{Best results for our DN-DETR and other detection models with the ResNet-50 backbone. $^*$ indicates it is the test-dev result.
    }
    \resizebox{1\textwidth}{!}{%
    \begin{tabular}{lccccccccccc}
        \toprule
        Model & MultiScale & \#epochs & AP & AP$_{50}$ & AP$_{75}$ & AP$_{S}$ & AP$_{M}$ & AP$_{L}$ & GFLOPs & Params \\
        \midrule
        Deformable DETR-R$50$ \cite{zhu2020deformable}  & \checkmark & $50$ & $43.8$ & $62.6$ & $47.7$ & $26.4$ & $47.1$ & $58.0$  & $173$ & $40$M \\ 
        SMCA-R$50$  \cite{gao2021fast}              & \checkmark & $50$ &  $43.7$ &  $63.6$ &  $47.2$ &  $24.2$ &  $47.0$ &  $60.4$ & $152$ & $40$M \\
        TSP-RCNN-R$50$ \cite{sun2020rethinking}         & \checkmark & $96$ & $45.0$ & $64.5$ & $49.6$ & $29.7$ & $47.7$ & $58.0$ & $188$ & $-$ \\
        Dynamic DETR-R$50^*$ \cite{dai2021dynamic} & \checkmark & $50$ & ${47.2}$ & $65.9$ & $51.1$ & $28.6$ & $49.3$ & $59.1$ & $-$ & $-$ \\
        DAB-Deformable-DETR-R$50$    &\checkmark & $50$ & {$46.9$} & $66.0$ & $50.8$ & $30.1$ & $50.4$ & $62.5$ & $195$ & $48$M \\
        % DAB-DETR-DC5-R$50$    & & $50$ & {$45.7$} & $66.2$ & $49.0$ & $26.1$ & $49.4$ & $63.1$ & $216$ & $44$M \\
        % DN-DETR-DC5-R$50$    & & $50$ & $\textbf{46.8}$ & $66.9$ & $50.1$ & $27.2$ & $50.6$ & $64.7$ & $--$ & $--$M \\
        DN-Deformable-DETR-R$50$    & \checkmark&50 &\textbf{48.6}& $67.4$ & $52.7$ & $31.0$ & $52.0$ & $63.7$ & $195$ & $48$M\\
        DN-Deformable-DETR-R$50$++    & \checkmark&50 &\textbf{49.5}& $67.6$ & $53.8$ & $31.3$ & $52.6$ & $65.4$ & $-$ & $47$M\\

        \bottomrule
    \end{tabular}}
    
    \centering

    % \vspace{-0.4cm}
    \label{tab:sota}
    % \vspace{-.3cm}
\end{table*}
We adopt several ResNet models ~\cite{he2015deep} pre-trained on ImageNet as our backbones and report our results on $4$ ResNet settings:  ResNet-50 (R50), ResNet-101 (R101), and their 16×-resolution extensions ResNet-50-DC5 (DC5-R50) and ResNet-101-DC5 (DC5-R101). For hyperparameters, we follow DAB-DETR to use a $6$-layer Transformer encoder and a $6$-layer Transformer decoder and $256$ as the hidden dimension. We add uniform noise on boxes and set the hyperparameters with respect to noise as $\lambda_1=0.4$, $\lambda_2=0.4$, and $\gamma=0.2$. For the learning rate scheduler, we use an initial learning rate (lr) $1\times 10^{-4}$ and drop lr at the 40-th epoch by multiplying 0.1 for the 50-epoch setting and at the 11-th epoch by multiplying 0.1 for the 12-epoch setting. We use the AdamW optimizer with weight decay of $1\times 10^{-4}$ and train our model on 8 Nvidia A100 GPUs. The batch size is 16. Unless otherwise specified, we use 5 denoising groups.
\par
We conduct a series of experiments to demonstrate the performance improvement as shown in Table \ref{tab:absoulute_results}, where we follow the basic settings in DAB-DETR without any bells and whistles in training. To compare with the state-of-the-art performance in the 12 epoch setting (the so-called $1\times$ setting in Detectron2) and the standard 50 epoch setting (most widely used in DETR-like models) in Table \ref{tab:1x} and \ref{tab:sota}, we follow DAB-DETR to use 3 pattern embeddings as in Anchor DETR ~\cite{wang2021anchor}. All our comparisons with DAB-DETR and its variants are under exactly the same setting.
\\
\textbf{DN-Deformable-DETR and DN-Deformable-DETR++:} For {DN-Deformable-DETR} with only deformable encoder, we use $10$ denoising groups. For {DN-Deformable-DETR++} with deformable attention in both encoder and decoder, we use $5$ denoising groups.  Note that we strictly follow Deformable DETR to use multi-scale ($4$ scale) features without FPN. Dynamic DETR~\cite{dai2021dynamic} adds FPN and more scales ($5$ scales) which can further boost the performance, but our performance still exceeds theirs.\\
\textbf{Faster R-CNN and Anchor DETR: } We use $10$ and $5$ denoising groups respectively.\\
\textbf{DINO: } To test the effectiveness of denoising training in DINO, we only use our proposed DN without its proposed contrastive DN and keep all the other components in DINO. We use $5$ denoising groups.\\
\textbf{Mask DINO: } Mask DINO incorporates both box denoising and mask denoising. To show performance improvement over segmentation tasks, we keep the box denoising part and only remove the mask denoising to study its effectiveness. We use $5$ denoising groups under this setting.
\\\textbf{Mask2Former: } Mask2Former is only designed for segmentation tasks. Therefore, we only add mask denoising training in our experiments. We use $5$ denoising groups under this setting.

Our proposed denoising training has been incorporated into many subsequent works and also implemented in detrex (\url{https://github.com/IDEA-Research/detrex}).

\subsection{Denoising Training Improves Performance}
To show the absolute performance improvement compared with DAB-DETR and other single-scale DETR models, we conduct a series of experiments using different backbones under the basic single-scale settings. The results are summarized in Table \ref{tab:absoulute_results}. \par
The results show that we achieve the best results among single-scale models with all four commonly used backbones. For example, compared with our baseline DAB-DETR under exactly the same setting, we achieve \textbf{+1.9} AP absolute improvement with ResNet-50. The table also shows that denoising training adds negligible parameters and computation.
\begin{figure*}
    \centering
    \begin{center}
\includegraphics[width=.60\columnwidth]{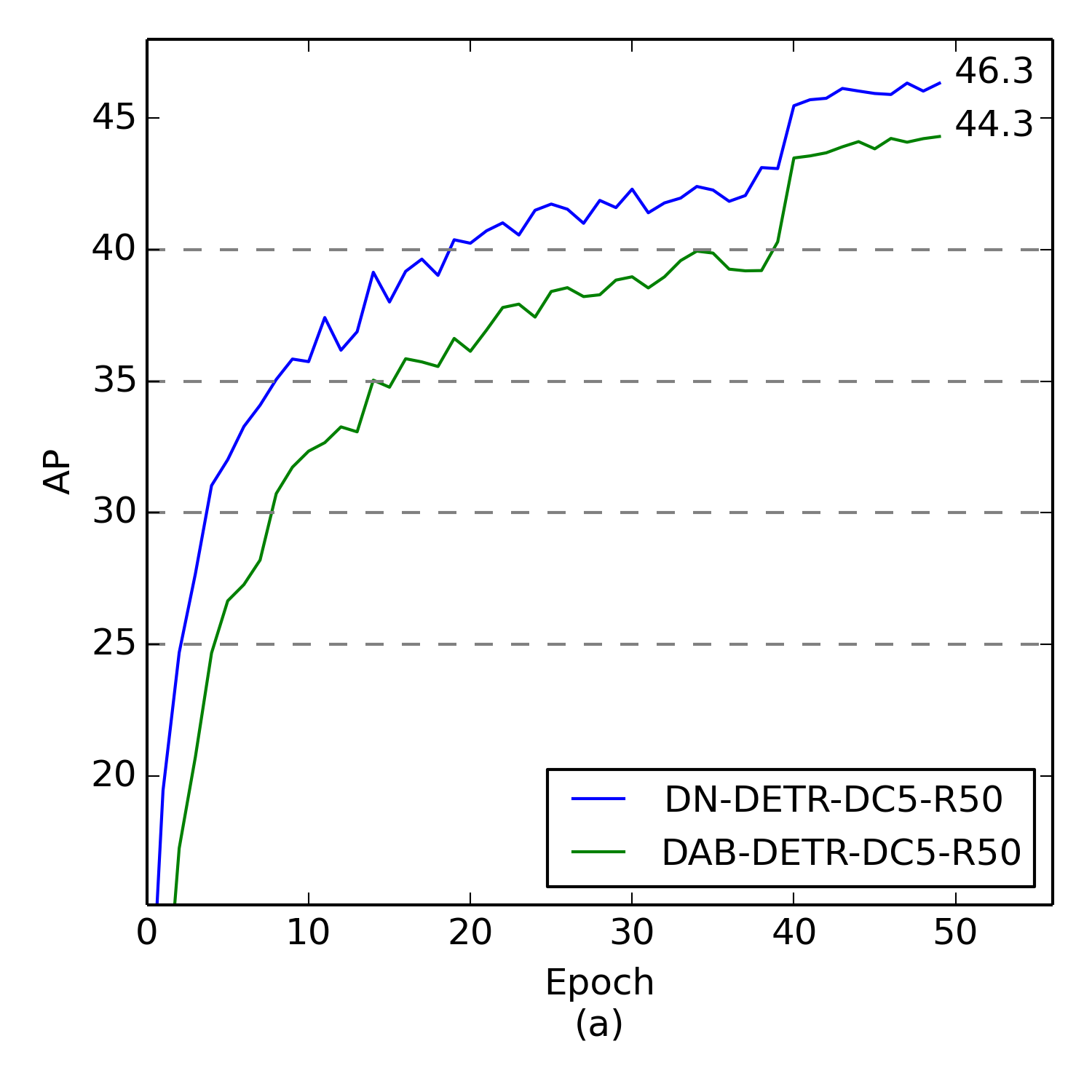}
% \subcaption[a]{}
% \caption{Convergence curve bettween DAB-DETR and DN-DETR on ResNet-DC5-50.}
% \hfill
\vspace{-0.1cm}
\includegraphics[width=1.08\columnwidth]{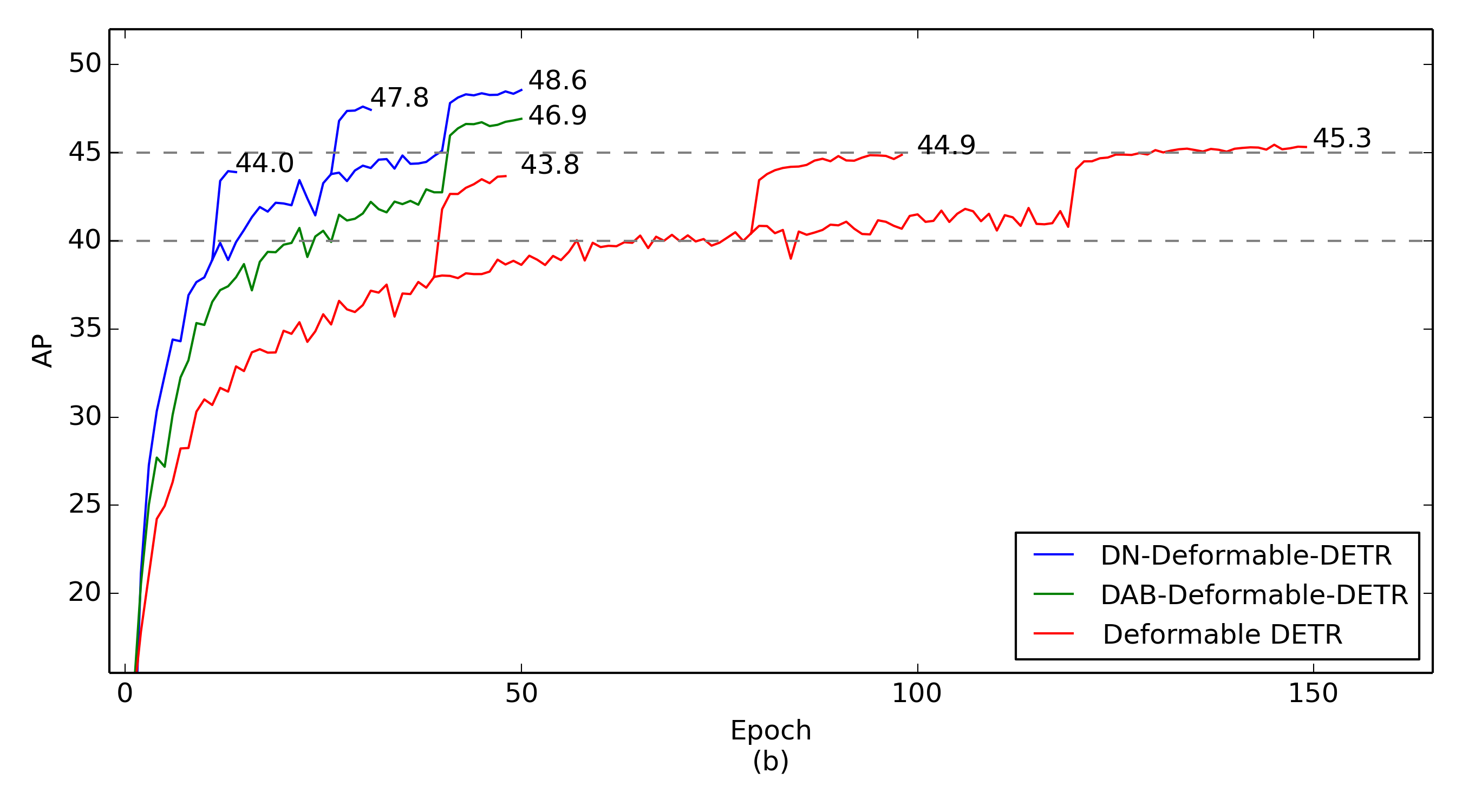}
% \vspace{-0.3cm}
\end{center}
    \caption{(a) Convergence curves of DAB-DETR and DN-DETR with ResNet-DC5-50. Before learning rate drop, DN-DETR achieves $40$ AP in $20$ epochs, while DAB-DETR needs $40$ epochs.  (b) Convergence curves of multi-scale models
    % Deformable DETR, DAB-Deformable-DETR, and DN-Deformable-DETR
    with ResNet-$50$. With learning rate drop, DN-Deformable-DETR achieves $47.8$ AP in 30 epochs, which is $0.9$ AP higher than the converged DAB-Deformable-DETR.}
    \label{fig:convergence_curve}
    % \vspace{-0.3cm}
\end{figure*}

\subsection{$1\times$  Setting}\label{sec:multi-scale}
With denoising training, the detection task can be accelerated by a large margin. 
As shown in Table \ref{tab:1x}, we compare our method with both a traditional detector ~\cite{ren2015faster} and some DETR-like models, including DETR~\cite{carion2020end}, Dynamic DETR~\cite{dai2021dynamic}, and Deformable DETR~\cite{zhu2020deformable}. Note that Dynamic DETR ~\cite{dai2021dynamic} adopts a dynamic encoder, for a fair comparison, we also compare with its version without a dynamic encoder.
\par
Under the same setting with the DC5-R50 backbone, DN-DETR can outperform DAB-DETR by $\textbf{+3.7}$ AP within $12$ epochs. Compared with other models, DN-Deformable-DETR achieves the best results in the $12$ epoch setting. It is worth noting that our DN-Deformable-DETR achieves $\textbf{44.1}$ AP within 12 epochs with the ResNet-101 backbone, which surpasses Faster R-CNN ResNet-101 trained for $108$ epochs ($9\times$ faster). 

\subsection{Extending DN to Other Detection and Segmentation Models}
To further validate the effectiveness of denoising training, we extend this method to other detection and segmentation model, as shown in Table \ref{tab:dn_series}. The experimental results indicate that denoising training is a universal training method to boost performance.

For example, we improve the DETR-like detection models significantly by $1.2-2.6$ AP under the 12-epoch setting. The results also reveal that
\begin{itemize}
    \item Denoising training is compatible with other positional query formulations, for example, Vallina DETR with high dimensional vectors, Anchor DETR with 2D anchor points, and DAB-DETR with 4D anchor boxes.
    \item Our method is only a training method and also compatible with other methods, for example, deformable attention \cite{zhu2020deformable}, semantic-alignment \cite{zhang2022semantic}, and query selection\cite{zhang2022dino}, etc.
\end{itemize}
\subsection{Compared with State-of-Art Detectors}

We also conduct experiments to compare our method with multi-scale models. The results is summarized in Table \ref{tab:sota}.
Our proposed DN-Deformable-DETR achieves the best result $\textbf{48.6}$ AP with the ResNet-50 backbone. 
To eliminate the performance improvement from formulating the queries of deformable DETR as anchor boxes, we further use a strong baseline DAB-Deformable-DETR without denoising training. 
The results show that we can still yield $1.7$ AP absolute improvement. The performance improvement of DN-Deformable-DETR also indicates that denoising training can be integrated into other DETR-like models and improve their performance. Though it is not a fair comparison with Dynamic DETR as it includes a dynamic encoder and more scales ($5$ scales) with FPN, we still yield $+1.4$ AP improvement. \par
We also show the convergence curve in both single-scale and multi-scale settings in Fig. \ref{fig:convergence_curve}, where we drop the learning rate by $0.1$ in multiple epochs in Fig. \ref{fig:convergence_curve}(b). 
% The detailed training acceleration analysis and training efficiency is shown is Appendix \ref{sec:accelartion_analysis} and \ref{sec:training_time}.

\subsection{Ablation Study}
\label{ablation}
\begin{table}[t]
    \centering
        \caption{Ablation results for DN-DETR. All models are trained with the ResNet-50 backbone using 1 denoising group under the same default settings.
    }
        \footnotesize
            \renewcommand{\arraystretch}{1.3}
    \resizebox{0.48\textwidth}{!}{%
    \begin{tabular}{cccc}
        \hline
        Box Denoising& Label Denoising & Attention Mask   & AP \\
        \hline
       
       \checkmark & \checkmark & \checkmark &43.4 \\
         \checkmark &  & \checkmark & 43.0 \\
        &  &\checkmark   & 42.2 \\
         \checkmark & \checkmark &    &  24.0\\
        % 5 & \checkmark& &\checkmark & \checkmark   &  41.9\\
        % % 6 & &\checkmark &\checkmark &    &  \\
        % 6 & & & &    & 39.5 \\
        \hline
    \end{tabular}}
    
    \label{tab:ablation}
    % \vspace{-.3cm}
\end{table}
\begin{table}[t]
    \centering
        \caption{Ablation results for DN-DETR using different numbers of denoising groups. All models are trained with the ResNet-50 backbone under the same default setting.
    }
        \footnotesize
            \renewcommand{\arraystretch}{1.3}
    \resizebox{0.4\textwidth}{!}{%
    \begin{tabular}{ccccccc}
        \hline
         &No Group&1 Group& 5 Groups  \\
        \hline
        R50 &42.2& 43.4 & 44.1  \\
        R50-DC5 & 44.5 & 45.6 & 46.3 \\
        R101 & 43.5 & 45.0 &45.2  \\
        R101-DC5 & 45.8 & 46.5 &47.3 \\
        
        \hline
    \end{tabular}}
    
    % \vspace{-0.5cm}
    \label{tab:ablation_head}
    % \vspace{-.3cm}
\end{table}
\subsubsection{Effectiveness of each component}
We conduct a series of ablation studies with the ResNet-50 backbone trained for 50 epochs to verify the effectiveness of each component and report the results in Table \ref{tab:ablation} and Table \ref{tab:ablation_head}. The results in Table \ref{tab:ablation} show that each component in denoising training contributes to performance improvement. Notably, without an attention mask to prevent information leakage, the performance degenerates significantly.
\subsubsection{Effectiveness of using more denoising groups}
We also analyze the influence of the number of denoising groups in our model, as shown in Table \ref{tab:ablation_head}. The results indicate that adding more denoising groups improves performance, but the performance improvement becomes marginal as the number of denoising groups increases. Therefore, in our experiment, our default setting uses 5 denoising groups, but more denoising groups can further boost performance as well as faster convergence.
\par
\begin{figure}[htbp]
\centering
\includegraphics[width=\linewidth]{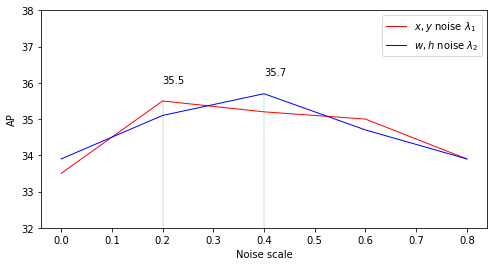}
\centering
\caption{DN-DETR in different noise scales. We fix one noise scale to $0.4$ and change the other. Noise scale is defined in \ref{denoising}}
\label{fig:noise ab}
\end{figure}

In Fig. \ref{fig:noise ab}, We explore the influence of noise scale. We run $20$ epochs with batch size $64$ and ResNet-50 backbone without learning rate drop. The results show that both center shifting and box scaling improve performance. But when the noise is too large, the performance drops.
\setcounter{table}{0}
\begin{table*}
    \centering
    \caption{Results of our method trained for $25$ epochs and our baseline method trained for $50$ epochs under the same settings. The results show we achieve $2$x acceleration with denoising training.
    }
    \resizebox{1\textwidth}{!}{%
    \begin{tabular}{lccccccccccc}
        \toprule
        Model & MultiScale & \#epochs & AP & AP$_{50}$ & AP$_{75}$ & AP$_{S}$ & AP$_{M}$ & AP$_{L}$ & GFLOPs & Params \\
        \midrule
        DAB-DETR-DC$5$-R$50$    & & $50$ & {${44.5}$} & $65.1$ & $47.7$ & $25.3$ & $48.2$ & $62.3$ & $202$ & $44$M \\
        DN-DETR-DC$5$-R$50$    & &25 &$44.4$& $64.5$ & $47.3$ & $24.4$ & $48.0$ & $63.0$ & $202$ & $44$M\\
        \hline
        DAB-Deformable-DETR-R$50$    &\checkmark & $50$ & {$46.9$} & $66.0$ & $50.8$ & $30.1$ & $50.4$ & $62.5$ & $195$ & $48$M \\
        DN-Deformable-DETR-R$50$    & \checkmark&25 &$46.8$& $65.5$ & $50.8$ & $28.9$ & $50.2$ & $62.5$ & $195$ & $48$M\\
        \hline
        DAB-Deformable-DETR-R$50$++    &\checkmark & $50$ & {$48.7$} & $67.2$ & $53.0$ & $31.4$ & $51.6$ & $63.9$ & $-$ & $47$M \\
        DN-Deformable-DETR-R$50$++    & \checkmark&25 &$48.4$& $66.6$ & $52.7$ & $30.0$ & $51.7$ & $64.4$ & $-$ & $47$M\\
        \hline
        Vanilla-DETR-R$50$ \cite{carion2020end}&&$500$ & $42.0$ & $62.4$ & $44.2$ & $20.5$ & $45.8$ & $61.1$ & $86$ & $41$M\\
        DN-Vanilla-DETR-R$50$&& $250$ & ${42.2}$ & $61.8$ & $44.6$ & $20.5$ & $46.0$ & $61.3$ &$86$ & $37$M\\
        \bottomrule
    \end{tabular}}
    
    \centering
    \label{tab:speedup}
    % \vspace{-.3cm}
\end{table*}
\subsubsection{Acceleration Analysis}\label{sec:accelartion_analysis}
We show how much our method can speed up training exactly in Table \ref{tab:speedup}. Our method achieves results comparable to the baseline with only  half of the training epochs, resulting in $2$x acceleration.

\subsubsection{The training wall clock time and GFLOPs}\label{sec:training_time}
We tested the training wall clock time and GFLOPs with 8 NVIDIA A100 GPUs as shown in Table~\ref{tab:training time}.
\begin{table}[ht]
   
        \footnotesize
            \renewcommand{\arraystretch}{1.3}
        \caption{We adopted five denoising groups for DN-DAB-DETR. The results are tested on the same GPUs for a fair comparison.
    }
    % \centering
    \begin{tabular}{ccc}
        \toprule
        Model  & Total Training time (min) & Training GFLOPs   \\
        \midrule
        DAB-DETR-R$50$    & $2555$($50$ epochs) & {${94.4}$}   \\
        DN-DAB-DETR-R$50$    &$1443$($25$ epochs) &$94.5$ \\
        
        \bottomrule
    \end{tabular}
    % \centering
    \label{tab:training time}
\end{table}
The total training time is calculated by multiplying the number of training epochs and the training time for each epoch. The training time per epoch is $51.1$min and $57.7$min for DAB-DETR-R50 and DN-DAB-DETR-R50, respectively. While denoising training introduces a minor training cost increase, it only needs about half the number of training epochs ($25$ epochs) to achieve the same performance as DAB-DETR-R50.
% Therefore, the total training times are $2555$ and $1443$ mins for DAB-DETR-R50 and DN-DAB-DETR-R50, respectively. 
The practical training speedup is indeed remarkable.
\subsection{Other tasks and future work}
\subsubsection{Other Tasks}
% We demonstrate some potential abilities of our method and some attempts we have made.
 In addition to regular detection, our design of queries as anchor box + label makes the detection model capable of handling other tasks. For example, known object detection and known label detection. Note that the results shown in this section are just a preliminary exploration and not based on our well-trained model with the best hyper-parameters.

\noindent\textbf{Known Object Detection: } Assume we know a part of the objects in an image and want to predict the remaining objects. We want the known objects to help predict the unknown objects through co-occurrence relations. We did some preliminary exploration. We randomly divide the $80$ classes of MS COCO2017 into $2$ parts, including known classes and unknown classes. We put objects of known classes in the denoising part and want the matching part to predict the objects of the unknown classes. We do not use an attention mask so that the matching part can get useful information from the denoising part. Our experimental results are shown in Table \ref{tab:extra_label}. Compared with the evaluation without known boxes, the evaluation of the known object improves the performance, which indicates that co-occurrence helps the prediction of unknown boxes. Moreover, our DN-DETR trained with known objects exceeds DAB-DETR only trained on unknown classes when evaluating without known objects. This means the denoising of extra boxes from extra (known) classes also helps the performance of the unknown objects.
\begin{table}[ht!]
  \centering
  \caption{Extra label prediction on COCO. We split the annotation of COCO class  into known/unknown classes, where objects of known classes only appear in denoising part, and we evaluate the performance on the unknown classes. Cond means the result is evaluated with known objects.}
  \resizebox{0.4\textwidth}{!}{%
  \begin{tabular}{@{}lclc@{}}
    \toprule
    Method & Setting & AP  & AP(Cond)\\
    \midrule
    DAB-DETR &$0.7/0.3$&$38.4$&-\\
    DN-DETR &$0.7/0.3$&$42.1$&$42.9$\\
    \midrule
    DAB-DETR & $0.5/0.5$&$37.8$&-\\
    DN-DETR & $0.5/0.5$&$39.1$&$40.3$\\
    \bottomrule
  \end{tabular}}
  \label{tab:extra_label}
\end{table}

\noindent\textbf{Known Label Detection: } For each image, we assume we know all the class labels in the image without box information. Since our model has interpreted the query embedding into class label embedding, we can seamlessly utilize these known labels to detect the boxes of each class label. For each class $c$ in the image, we concatenate its label embedding with the indicator $1$, which denotes a known label. We feed the concatenated vector into the decoder and let the decoder output all boxes of class $c$. To compare with methods without known labels and detect all objects in an image, we concatenate outputs of all classes and evaluate the result as shown in Table \ref{tab:known_label}.
By finetuning with known labels, the detection performance can be improved in only one epoch. Within $10$ epochs of finetuning on pre-trained DN-DETR, the known label detection performance is improved to $46.6$. This result demonstrates that given labels can significantly improve the detection performance. 
\begin{table}[ht!]
  \centering
    \caption{Known label detection results under ResNet-50 with $1$ denoising group. $1$ep and $10$ep means finetuned $1$ or $10$ epochs from pretrained DN-DETR.}
    \resizebox{0.4\textwidth}{!}{%
  \begin{tabular}{@{}lcl@{}}
    \toprule
    Method & Setting & AP  \\
    \midrule
    DAB-DETR &no knwon labels&42.2\\
    DN-DETR & no knwon labels&43.4 \\
    DN-DETR &known label ($1$ep)&43.8 \\
    DN-DETR & known label ($10$ep)&46.6 \\
    \bottomrule
  \end{tabular}}
  \label{tab:known_label}
\end{table}
\subsubsection{Future Work}
There are three potential future works to be mentioned here. One is zero-shot detection, and the other is progressive inference.\\
\textbf{Zero-shot or Open Set Detection: } Since we have decoupled decoder queries as anchor boxes and class labels, pre-trained class label embeddings can be fed into the class label part of the queries. To enable zero-shot detection, one can take $80$ classes of MSCOCO as phrases and collect phrase embeddings from a pre-trained language model as the class label embedding. With the pre-trained label embedding, it is possible to train a given class detector that takes a class label embedding as input and detects objects of the given classes. In inference time, class label embeddings from unseen classes can be fed into the decoder to achieve zero-shot detection.

\noindent\textbf{Progressive inference: }
Based on known object detection, a progressive inference method can be designed. For example, we can train a DN-DETR capable of doing known object detection. In inference time, we let the detector predict objects, and then, we can choose the objects with the highest score and treat them as known objects to do known object detection. For each step of prediction, we choose objects with the highest score and add them to the known box set. After repeating for many times, we get the final prediction.

\noindent\textbf{Classification before detection: }
As shown in Table~\ref{tab:known_label}, given labels can significantly improve the detection performance. Therefore, one potential future work is to add a muli-label classification network to provide labels and feed them to DN-DETR, which may help improve detection performance.
\section{Conclusion}
In this paper, we have analyzed the reason for the slow convergence of DETR training lying in the unstable bipartite matching and proposed a novel denoising training method to address this problem. Based on this analysis, we proposed DN-DETR by integrating denoising training into DAB-DETR to test its effectiveness. DN-DETR specifies the decoder embedding as label embedding and introduces denoising training for both boxes and labels. We also added denoising training to Deformable DETR to show its generality. The results show that denoising training significantly accelerates convergence and improves performance, leading to the best results in the 1x (12 epochs) setting with both ResNet-50 and ResNet-101 as the backbone. This study shows that denoising training can be easily integrated into DETR-like models as a general training method with only a small training cost overhead and bring in a remarkable improvement in terms of both training convergence and detection performance.\\
\textbf{Limitations: } 
In this work, the added noises are simply sampled from a uniform distribution. We have not explored more complex noising schemes and leave these for future work. 
Reconstructing noised data achieves great success in unsupervised learning and diffusion models. This work is an initial step to apply it to object detection. In the future, we will explore how to pre-train detectors on weakly labeled data with unsupervised learning techniques and explore applying other denoising training schemes in detection models. 

{\small
\bibliographystyle{ieee_fullname}
\bibliography{egbib}
}
% \clearpage
% \include{cvpr2022/SUP}

\end{document}